\documentclass[11pt]{article}

\usepackage[final]{acl}

\usepackage{times}
\usepackage{latexsym}

\usepackage[T1]{fontenc}

\usepackage[utf8]{inputenc}

\usepackage{microtype}

\usepackage{inconsolata}

\usepackage{graphicx}
\usepackage{amsmath}
\usepackage{booktabs}
\usepackage{array}
\usepackage{xcolor}
\usepackage{enumitem}
\usepackage{tikz}
\usetikzlibrary{positioning, arrows.meta, shapes.geometric, fit}
\title{\textsc{REDACT}: A Systematically Controlled Multilingual Benchmark for Personal Information Detection}

\author{
\textbf{Guneesh Vats} \quad
\textbf{Anubha Agrawal} \quad
\textbf{Shikha Singhal} \quad
\textbf{Ajita Dash} \\
\textbf{Praison Selvaraj} \quad
\textbf{Vidhan Jhawar} \quad
\textbf{Ranga Prasad Chenna} \quad
\textbf{Bharadwaj Y M G} \\
ServiceNow \\
\texttt{\{guneesh.vats, anubha.agrawal, shikha.singhal, ajita.dash\}@servicenow.com} \\
\texttt{\{praison.selvaraj, vidhan.jhawar, rangaprasad.chenna, bharadwaj.ymg\}@servicenow.com}
}

\begin{document}
\maketitle
\begin{abstract}
Benchmark infrastructure for personally identifiable information
(PII) detection remains limited: existing corpora cover few entity
types, use ad hoc generation conditions, and do not show which
surface conditions cause detector failures. We present \textsc{REDACT}, a systematically controlled multilingual
PII benchmark with $13{,}427$ records, $324{,}078$ entity annotations,
$51$ entity types, $4{,}127$ surface-form patterns, and $25$ languages
across $9$ scripts. A strength-2 covering-array sampler controls nine
generation axes: domain, format, difficulty, length, density,
code-switching, language, adjacency, and co-occurrence. Three
entity-level metadata fields (disclosure status, disclosure form, and
a GDPR-aligned sensitivity tier) enable stratified evaluation beyond
aggregate or per-type F1. From the full benchmark, we evaluate five
detectors (Presidio, GLiNER, the OpenAI Privacy Filter, GPT-4.1, and
Claude Sonnet~4.6) on a locked, language-stratified sample of
$1{,}000$ records. Aggregate F1 masks an architecture-dependent
failure structure: the rule-based detector performs poorly on the
highest-stakes data, including HIGH-sensitivity categories (recall
$0.07$) and non-verbatim disclosure forms, while the LLM detectors
remain more robust, with the HIGH tier as their strongest sensitivity slice. A
three-model reference-free LLM-as-judge assessment corroborates
that sensitivity-tier assignment is the task's hardest axis. We release
the benchmark, schema, prompts, and stratified evaluation harness.
\end{abstract}

\section{Introduction}
\label{sec:introduction}

Personally identifiable information (PII) detection in unstructured
text is governed by data-protection regulation: the EU's General Data
Protection Regulation \citep{gdpr2016}, the U.S.\ Health Insurance
Portability and Accountability Act, and equivalent regimes in Brazil
and Japan require organizations to identify and process identifiers
from logs, chat transcripts, and customer-service records with
documented accuracy. A detector that catches names but misses national
identification numbers, handles English but not Hindi, or recognizes
\textit{John Smith} but not \textit{J.S.}\ or \textit{[REDACTED]}
falls short along axes current benchmarks do not measure.

Three gaps in the current evaluation landscape motivate this work.
First, existing PII benchmarks operate in narrow multilingual
regimes: TAB \citep{pilan2022tab} and the i2b2/n2c2 challenges
\citep{stubbs2017identifying} cover only English; SPY
\citep{savkin2025spy} covers two English domains; AI4Privacy
\citep{ai4privacy2025openpii} spans $23$ languages but uses templated
generation that limits behavioral diversity; and PIIBench
\citep{jha2026piibench} consolidates ten English sources into a
unified taxonomy. This leaves practitioners with little evidence about
how detectors behave under the multilingual, mixed-format, and
compliance-sensitive conditions seen in deployment. Second, aggregate
F1 is the dominant metric, yet high scores routinely mask severe
per-entity failures \citep{fu2020interpretable}; detectors can perform
unevenly across entity types once the taxonomy is broad enough to
expose those differences \citep{jha2026piibench}. What is missing is
a systematic way to surface such failures along regulatorily meaningful
axes. Third, no public PII benchmark records whether a mention is
\emph{actually disclosed} versus hypothetical, redacted, or only
partially exposed, even though these distinctions determine real-world
detector utility.

We introduce \textsc{REDACT}, a multilingual PII detection benchmark constructed to address these three gaps.\footnote{All code, the generation,
detector, and judge prompts, the quality rubric, and the evaluation
harness are released at the anonymized repository:
\url{https://github.com/guneeshvats/REDACT-PII-Benchmark}.} Our contributions are:
\textbf{(1)} a $51$-type PII taxonomy across $25$ languages with an explicit HIGH-sensitivity slice covering GDPR Article~9 special-category data, Article~10-style criminal-offence data, and linkage-risk identifiers (Section~\ref{sec:taxonomy}); \textbf{(2)} a nine-axis
stratified sampler that exercises every pair of axis values via a
strength-2 covering array \citep{kuhn2003covering}
(Section~\ref{sec:sampler}); \textbf{(3)} three entity-level
metadata fields (\texttt{disclosed}, \texttt{disclosure\_form},
\texttt{sensitivity\_tier}), inspired by privacy-oriented
anonymization annotation in TAB \citep{pilan2022tab}, powering a
stratified-evaluation framework
(\textbf{F5} per disclosure form, \textbf{F6} per sensitivity tier)
that reveals detector failures aggregate F1 hides
(Section~\ref{sec:metadata}); \textbf{(4)} a reproducible seven-stage
generation pipeline with deterministic offset alignment, conditional
verification, and a six-metric release-gate protocol
(Section~\ref{sec:pipeline}); and \textbf{(5)} an empirical evaluation
of five detectors (Presidio \citep{presidio2024}, GLiNER
\citep{zaratiana2024gliner}, the OpenAI Privacy Filter
\citep{openai2026privacyfilter}, GPT-4.1, Claude Sonnet~4.6) under
partial, exact, and fuzzy span matching (Section~\ref{sec:eval}).

\section{Related Work}
\label{sec:related}

\textbf{PII benchmarks.} Existing PII resources make different
coverage trade-offs. Real-text corpora such as the i2b2/n2c2
challenges \citep{stubbs2017identifying} and TAB
\citep{pilan2022tab} provide carefully annotated data, but they are
limited to English and to a small number of domains. Synthetic and
consolidated resources broaden the setting: AI4Privacy spans many
languages \citep{ai4privacy2025openpii}, SPY studies synthetic PII in
two English domains \citep{savkin2025spy}, and PIIBench consolidates
English sources into a larger taxonomy \citep{jha2026piibench}. These
benchmarks are useful, but they do not jointly provide broad
multilingual coverage, fine-grained PII types, disclosure-form labels,
and sensitivity-tier slices. Our benchmark is designed to fill this
combined evaluation gap (Appendix~\ref{sec:relatedtable}).
TAB also motivates treating privacy annotation as more than span
classification, since anonymization decisions depend on whether and
how information should be masked. We build on that direction by
recording disclosure status, disclosure form, and sensitivity tier at
the entity level, then using those fields for stratified recall.
Detector-side systems such as GLiNER2-PII
\citep{hurnmaloney2026gliner2pii} and RECAP
\citep{rajgarhia2025recap}, as well as model-level privacy benchmarks
such as PrivLM-Bench \citep{li2024privlm}, are complementary to this
work because they can be evaluated using the substrate we provide.

\textbf{Evaluation methodology.} Our stratified metrics build on
attribute-based NER evaluation \citep{fu2020interpretable} and
behavioral testing \citep{ribeiro2020checklist}, applying them to PII
attributes that matter for compliance. For construction, we draw on
single-call data generation \citep{xu2024magpie}, iterative
deduplication \citep{wang2023selfinstruct}, and prior multilingual NER
benchmarks \citep{malmasi2022multiconer,fetahu2023multiconer2} to set
language and coverage targets.

\section{Dataset Construction}
\label{sec:construction}

We construct the benchmark to satisfy three operational requirements
that prior PII corpora address only partially: (i) coverage across a
broad multilingual taxonomy aligned with high-sensitivity privacy categories, (ii)
stratification along axes that surface detector failure modes hidden
by aggregate metrics, and (iii) reproducibility from a fixed catalog
of $4{,}127$ PII patterns. The construction proceeds through a seven-stage record-generation pipeline after taxonomy and pattern-catalog specification: stratified plan generation, single-call LLM generation, deterministic span checking and offset alignment, conditional verification, repair, deduplication, and quality auditing.

\subsection{Taxonomy}
\label{sec:taxonomy}

We define $51$ canonical entity types organized into seven families:
identification, contact, demographic, geographic, financial,
professional, and digital. All $51$ are listed in the per-type results
(Appendix~\ref{sec:pertype-app}, Figure~\ref{fig:pertype}), with the
HIGH-sensitivity categories banded there. The 51-type granularity is finer
than the approximately ten categories of TAB \citep{pilan2022tab} and
broader than the 30-category schema of \citet{jha2026piibench}. We treat the HIGH-sensitivity types as a distinguished evaluation slice in
Section~\ref{sec:metadata}.

\subsection{Pattern Catalog}
\label{sec:catalog}

For each entity type and language we compile a catalog enumerating the
surface forms an entity can take, assembled as the union of four
annotators' independent enumerations with rare and long-tail forms
explicitly solicited (Appendix~\ref{sec:catalog-construction}):
$4{,}127$ patterns across $25$ languages in three resource tiers (six
high-, three mid-, and sixteen lower-resource languages; full list in
Appendix~\ref{sec:axes}).
Patterns carry country-specific identifier formats with checksum
specifications (Luhn, IBAN mod-97, national-ID schemes), seven
non-Gregorian calendar systems for \textit{Date\_of\_Birth}, and
locale-appropriate name distributions.

\subsection{Nine-Axis Stratified Sampler}
\label{sec:sampler}

We define nine axes that, in combination, govern the distribution
from which records are sampled: \textit{domain} ($12$ values),
\textit{format} ($7$ values), \textit{difficulty} ($3$ values),
\textit{length} ($4$ values), \textit{density} ($3$ values),
\textit{code\_switching} ($3$ values), \textit{language} ($25$
values), \textit{adjacency} ($3$ values), and
\textit{co\_occurrence\_pattern} ($15$ patterns plus a sentinel).
The full value set for each axis is enumerated in
Appendix~\ref{sec:axes}.
A naive full factorial over these axes would require $10{,}886{,}400$
records to realize every combination exactly once. We instead adopt a
strength-2 covering array construction \citep{kuhn2003covering,nist800142},
which guarantees by design that every \emph{pair} of axis values
co-occurs in at least one planned record while reducing the planned
sample by three orders of magnitude. After the quality filtering and
deduplication described in Section~\ref{sec:qa}, the released corpus
realizes $98.2\%$ of all pairwise axis-value combinations across the
$\binom{9}{2}=36$ axis pairs. The remaining gap reflects records
removed during filtering rather than missing combinations in the
sampling plan.

The sampler additionally supports a paired-sweep mode that holds
entity content fixed while a single axis varies, isolating that axis's
effect on detector behavior. We release this mode as part of the
harness and discuss the quantitative analysis in the Limitations.

\subsection{Generation Pipeline}
\label{sec:pipeline}

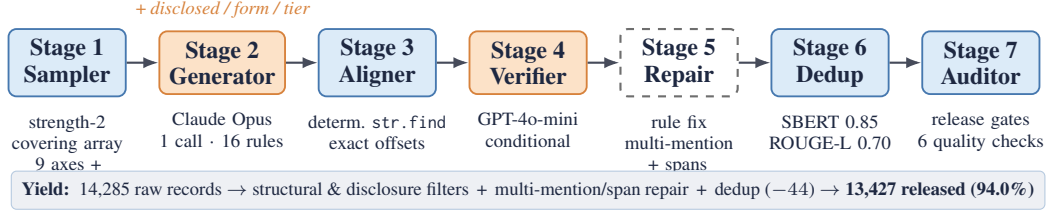
\begin{figure*}[t]
\centering
\definecolor{detfill}{RGB}{220, 235, 248}
\definecolor{detborder}{RGB}{70, 120, 175}
\definecolor{llmfill}{RGB}{252, 226, 200}
\definecolor{llmborder}{RGB}{215, 130, 60}
\definecolor{condborder}{RGB}{110, 110, 110}
\definecolor{arrcolor}{RGB}{70, 75, 90}
\definecolor{titletext}{RGB}{30, 40, 65}
\definecolor{yieldfill}{RGB}{238, 240, 244}
\begin{tikzpicture}[
    node distance=0.24cm and 0.42cm,
    every node/.style={font=\small},
    box/.style={rectangle, draw, rounded corners=3pt, line width=0.7pt,
        minimum width=1.55cm, minimum height=0.8cm, align=center,
        font=\small\bfseries, text=titletext},
    detbox/.style={box, fill=detfill, draw=detborder},
    llmbox/.style={box, fill=llmfill, draw=llmborder},
    condbox/.style={box, fill=white, draw=condborder, dashed, line width=0.8pt},
    arr/.style={-{Latex[length=2.0mm]}, thick, draw=arrcolor},
    sub/.style={font=\scriptsize, align=center, text width=1.9cm, text=titletext},
    note/.style={font=\scriptsize\itshape, align=center, text=llmborder, inner sep=1pt}
]
\node[detbox] (s1) {Stage 1\\Sampler};
\node[llmbox, right=of s1] (s2) {Stage 2\\Generator};
\node[detbox, right=of s2] (s3) {Stage 3\\Aligner};
\node[llmbox, right=of s3] (s4) {Stage 4\\Verifier};
\node[condbox, right=of s4] (s5) {Stage 5\\Repair};
\node[detbox, right=of s5] (s6) {Stage 6\\Dedup};
\node[detbox, right=of s6] (s7) {Stage 7\\Auditor};
\draw[arr] (s1) -- (s2); \draw[arr] (s2) -- (s3); \draw[arr] (s3) -- (s4);
\draw[arr] (s4) -- (s5); \draw[arr] (s5) -- (s6); \draw[arr] (s6) -- (s7);
\node[note, above=0.12cm of s2] {+\,disclosed / form / tier};
\node[sub, below=0.10cm of s1] {strength-2 covering array\\9 axes + paired sweep};
\node[sub, below=0.10cm of s2] {Claude Opus\\1 call $\cdot$ 16 rules};
\node[sub, below=0.10cm of s3] {determ.\ \texttt{str.find}\\exact offsets};
\node[sub, below=0.10cm of s4] {GPT-4o-mini\\conditional};
\node[sub, below=0.10cm of s5] {rule fix\\multi-mention + spans};
\node[sub, below=0.10cm of s6] {SBERT 0.85\\ROUGE-L 0.70};
\node[sub, below=0.10cm of s7] {release gates\\6 quality checks};
\node[below=1.0cm of s4, fill=yieldfill, rounded corners=2pt, draw=detborder!40,
      line width=0.4pt, inner sep=4pt, font=\scriptsize, text=titletext, align=center] (yield)
   {\textbf{Yield:}\ \ 14{,}285 raw records $\rightarrow$ structural \& disclosure filters \,+\, multi-mention/span repair \,+\, dedup ($-44$) $\rightarrow$ \textbf{13{,}427 released (94.0\%)}};
\end{tikzpicture}
\caption{Seven-stage generation pipeline. Deterministic stages compute the plan, offsets, deduplication, and release gates; LLM stages generate and verify records.}
\label{fig:pipeline}
\end{figure*}

Each plan row produced by the sampler is processed through a
seven-stage pipeline (Figure~\ref{fig:pipeline}) implemented atop the
SyGra synthetic data generation framework \citep{servicenow2025sygra}.
The \emph{generator} (Stage~2) issues a single LLM call per record
under a structured prompt of $16$ hard rules and $10$ few-shot anchors
(Appendix~\ref{sec:genprompt}); a single call rather than a multi-step
decomposition keeps all record-level constraints in one generation
context \citep{xu2024magpie}. The \emph{offset aligner} (Stage~3)
performs deterministic span checking by recomputing character offsets
with iterated \texttt{str.find}, so annotation integrity is checked
before release rather than left to post-hoc manual inspection. The
\emph{verifier} (Stage~4) repairs flagged records with GPT-4o-mini,
the \emph{deduplicator} (Stage~6) applies an SBERT cosine pre-filter
\citep{reimers2019sentence,gao2021simcse} followed by ROUGE-L
\citep{lin2004rouge,wang2023selfinstruct}, and the \emph{auditor}
(Stage~7) computes the release gates of Section~\ref{sec:qa}.

\subsection{Metadata Fields}
\label{sec:metadata}

The core annotation design adds three metadata fields to each
entity, beyond the standard type-and-span pair. These fields are
inspired by privacy-oriented annotation in TAB \citep{pilan2022tab},
but are operationalized here for multilingual stratified PII
evaluation.

\paragraph{disclosed.} A boolean separating entities actually
disclosed in the text from those mentioned hypothetically,
counterfactually, or in denial (e.g., ``\textit{if your accrual were
1.25 sick days per month}''), so that behavior on non-disclosed
mentions is scored rather than ignored.

\paragraph{disclosure\_form.} A field over \{\textit{complete},
\textit{partial}, \textit{obfuscated}\}: a fully spelled entity
(``\textit{John Smith}''), a truncated/initialed form (``\textit{J.
Smith}'', last four SSN digits), or class-preserving masking
(``\textit{REDACTED}'', ``\textit{[NAME]}''). Per-form recall reveals
degradation on the non-verbatim forms central to enterprise
deidentification.

\paragraph{sensitivity\_tier.} A field over \{\textit{HIGH},
\textit{MEDIUM}, \textit{LOW}\}. The HIGH tier includes GDPR
Article~9 special-category data, Article~10-style criminal-offence
data, and linkage-risk identifiers such as Date\_of\_Birth, which we
mark HIGH because of its identity and linkage risk. MEDIUM covers
direct identifiers outside this HIGH set, including names, account
numbers, emails, phones, addresses, and similar IDs; LOW covers
quasi-identifiers such as city, country of residence, and business
title. Appendix~\ref{sec:tier-rules} gives the decision rules used for
ambiguous cases such as nationality versus country of residence,
medical hints versus explicit medical information, and category
mentions. Per-tier recall isolates the high-sensitivity subset that
aggregate metrics dilute.

Validated by the auditor (Stage~7), these fields underpin the F5/F6
framework (Section~\ref{sec:eval}). The contribution is their
combined release and use as recall slices for multilingual PII
evaluation, rather than any single metadata label in isolation.

\subsection{Quality Assurance}
\label{sec:qa}

We apply a release-gate protocol of six critical quality metrics:
five for annotation integrity and one for coverage
(Appendix~\ref{sec:gates}). After structural filtering, multi-mention
and span repair, and near-duplicate removal (a further $44$ records),
the released corpus contains $13{,}427$ records, corresponding to a
$94.0\%$ retention rate (Figure~\ref{fig:pipeline}). The auditor checks six release gates; five pass, and B4 (triple-name decomposition) is the only failing gate. Appendix~B summarizes the gate definitions and pass/fail outcomes; the full auditor report is released with the harness.

\subsection{Corpus Statistics}
\label{sec:corpusstats}

The released corpus comprises $13{,}427$ records and $324{,}078$
entity annotations ($24.1$ per record). Coverage is balanced: each of
the $25$ languages contributes $453$--$621$ records across $9$
scripts, and each entity type appears in at least $50$ records. This
allows per-type evaluation for all $51$ types, including HIGH-sensitivity. Records with at least one HIGH-sensitivity entity total $3{,}829$ ($28.5\%$). The corpus instantiates $51.9\%$ of catalog patterns and realizes
$98.2\%$ of pairwise axis-value combinations; mean pairwise ROUGE-L is
$0.03$, confirming low residual near-duplication.

\section{Evaluation Framework}
\label{sec:eval}

We evaluate detectors on aggregate accuracy and on stratified recall
along the two metadata axes introduced in Section~\ref{sec:metadata}.

\subsection{Detector Baselines}
\label{sec:detectors}

We evaluate five detectors spanning the dominant deployment families.
\textbf{Microsoft Presidio} \citep{presidio2024} combines regular
expressions with spaCy NER \citep{honnibal2020spacy};
\textbf{GLiNER-multi} \citep{zaratiana2024gliner} is a zero-shot
multilingual span detector run with sliding-window inference; the
\textbf{OpenAI Privacy Filter} \citep{openai2026privacyfilter} is an
open-weight fine-tuned token classifier emitting eight coarse
categories mapped onto our taxonomy; and \textbf{GPT-4.1} and
\textbf{Claude Sonnet~4.6} are frontier LLMs used as zero-shot
extractors under identical instructions. All five run on a locked,
language-stratified sample of $1{,}000$ records from the full corpus
(Section~\ref{sec:statmethod}), with native schemas mapped to our
$51$-type taxonomy via the released catalogs.

\subsection{Span Matching}
\label{sec:matching}

We report micro- and macro-F1 under three modes: \emph{exact}
(coincident spans), \emph{partial} (any same-type overlap), and
\emph{fuzzy} ($\geq 50\%$ overlap of the shorter span). We use
partial-overlap as the primary metric, following TAB
\citep{pilan2022tab} and the de-identification literature
\citep{stubbs2017identifying}, where boundaries may vary even when
the detection decision is correct. LLM outputs require special
handling because the prompts ask models to return entity strings and
types, while character offsets are not always reliable. We therefore
relocate each LLM prediction using occurrence-tracked \texttt{str.find}
against the same input text. This step only maps the model's emitted
string back to a position; it does not change the predicted string or
entity type, and predictions that cannot be located are left unmatched.
Systems that emit native offsets are scored using those offsets. After
this normalization, all predictions are evaluated with the same exact,
partial, and fuzzy span matchers. This choice evaluates detection
quality separately from output-format artifacts while keeping strict
offset reliability visible through the exact-match setting.

\subsection{Stratified Recall}
\label{sec:stratified}

The framework's principal evaluation contribution is the
\emph{stratified} family, which decomposes aggregate performance
along the two metadata axes of Section~\ref{sec:metadata}. Because
disclosure form and sensitivity tier are properties of the
\emph{gold} entities, we stratify recall rather than F1: a
prediction's tier or form is defined only once it is matched to a gold
entity, so precision is not attributable to a gold slice.

\paragraph{F5: per-disclosure-form recall.} For each detector, we
compute recall separately on the subset of gold entities tagged
\textit{complete}, \textit{partial}, and \textit{obfuscated}. The
\emph{obfuscation gap}
$\Delta_{\text{F5}} = R_{\text{complete}} - R_{\text{obfuscated}}$
quantifies the degradation that aggregate F1 hides; a large positive
$\Delta_{\text{F5}}$ indicates that the detector relies on full
surface forms and cannot generalize to redacted or truncated
mentions.

\paragraph{F6: per-sensitivity-tier recall.} Analogously, we compute
recall stratified by \textit{HIGH}, \textit{MEDIUM}, and \textit{LOW}
sensitivity tiers. The \emph{HIGH-sensitivity gap}
$\Delta_{\text{F6}} = R_{\text{LOW}} - R_{\text{HIGH}}$ isolates
regulatory-sensitive performance. A positive $\Delta_{\text{F6}}$
indicates a detector that underperforms on the HIGH-sensitivity categories carrying the highest legal stakes; as Section~\ref{sec:results} shows, the \emph{sign} of this gap is determined by detector architecture.

We additionally report per-entity-type F1 (51 rows) and per-language
F1 (25 rows), extending PIIBench's per-type failure-mode analysis
\citep{jha2026piibench} to a larger taxonomy and language set.

\subsection{Prompts and Quality Assessment}
\label{sec:promptversion}\label{sec:geval}

Both detectors and judges use a final prompt revision with four-level
rubric anchors, bias controls, few-shot exemplars
\citep{brown2020language}, and an inline Article~9 mapping (all revisions are released with the harness). Beyond detector scoring we assess
intrinsic record quality with a reference-free LLM-as-judge protocol
after G-Eval \citep{liu2023geval}, scoring each record $1$--$10$ on six
weighted dimensions (realism, entity validity at $2.0$; type,
disclosed-flag at $1.5$; tier, span quality at $1.0$). The anchor
ranges used by the judges are listed in Appendix~\ref{sec:rubric}. A
three-family panel (GPT-5.2, Claude
Sonnet~4.6, Gemini~2.5~Pro) guards against single-vendor bias; Gemini
returned valid scores on $348$ of $500$ records and the others on all
$500$, each at its realized size, with inter-judge agreement by
pairwise Cohen's $\kappa$ over three ordinal bins.

\subsection{Sample Sizes and Statistical Power}
\label{sec:statmethod}

Evaluation uses three nested samples: the detector sample
($1{,}000$ records, $40$ per language, fixed seed), the quality panel
($500$ records), and manual validation ($510$ entities, $10$ per
type, two annotators, plus a $25$-record audit). For manual
validation, annotators independently checked each sampled entity
against the generated text for type, span, disclosed flag, disclosure
form, and sensitivity tier. The separate record-level audit checked
whether any PII mentions in the sampled records were missing from the
gold annotations. Per-language resolution is the binding constraint; Appendix~\ref{sec:power} gives the sizing rationale.

\section{Results}
\label{sec:results}

\subsection{Aggregate Detector Performance}
\label{sec:results-aggregate}

Table~\ref{tab:mainf1} reports aggregate performance on the locked
$1{,}000$-record sample. Under the primary partial-overlap metric,
Claude Sonnet~4.6 leads ($0.636$), followed by GPT-4.1 ($0.597$), the
OpenAI Privacy Filter ($0.512$), GLiNER ($0.320$), and Presidio
($0.195$). Macro-F1 and exact-overlap largely preserve this ranking;
Presidio's $0.063$ macro-F1 reflects near-zero recall on many types.

\begin{table}[t]
\centering
\small
\setlength{\tabcolsep}{4pt}
\begin{tabular}{>{\raggedright\arraybackslash}p{0.43\columnwidth} c c c}
\toprule
\textbf{Detector} & \textbf{Partial} & \textbf{Macro} & \textbf{Exact} \\
 & \textbf{micro-F1} & \textbf{F1} & \textbf{micro-F1} \\
\midrule
Presidio {\scriptsize\citep{presidio2024}} & 0.195 & 0.063 & 0.145 \\
GLiNER {\scriptsize\citep{zaratiana2024gliner}} & 0.320 & 0.224 & 0.293 \\
OpenAI Privacy Filter {\scriptsize\citep{openai2026privacyfilter}} & 0.512 & 0.254 & 0.171 \\
GPT-4.1                 & 0.597 & 0.565 & 0.558 \\
\textbf{Claude Sonnet~4.6} & \textbf{0.636} & \textbf{0.619} & \textbf{0.602} \\
\bottomrule
\end{tabular}
\caption{Aggregate detector performance on the locked
$1{,}000$-record sample. Partial-overlap micro-F1 is the
primary metric; macro-F1 averages over the $51$ canonical types;
exact-overlap micro-F1 requires coincident spans. LLM predictions
are matched after offset relocation (Section~\ref{sec:matching}).}
\label{tab:mainf1}
\end{table}

\subsection{Stratified Evaluation: an Architecture-Dependent Failure Structure}
\label{sec:results-stratified}

\begin{figure*}[t]
\centering
\includegraphics[width=0.92\textwidth]{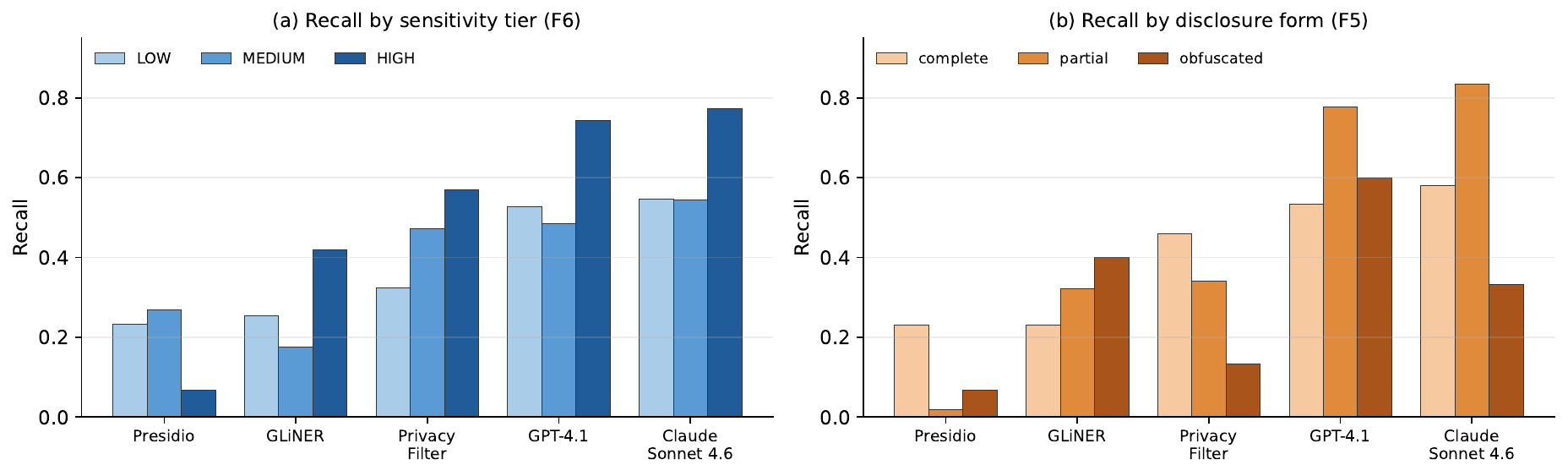}
\caption{Stratified recall by sensitivity tier and disclosure form. Presidio is weakest on the HIGH tier and on partial/obfuscated forms, while LLM detectors are strongest on the HIGH tier; the detailed HIGH-sensitivity and disclosure-form gaps are discussed in Section~\ref{sec:results-stratified}.}
\label{fig:stratified}
\end{figure*}

Figure~\ref{fig:stratified} shows the main value of the metadata
slices: detector failures are not uniform, but depend strongly on
architecture. On the sensitivity axis, Presidio drops from $0.23$
recall on LOW-tier entities to $0.07$ on the HIGH tier,
giving an HIGH-sensitivity gap of $\Delta_{\text{F6}}=+0.16$. The non-regex
systems show the opposite pattern: GLiNER, the OpenAI Privacy Filter,
GPT-4.1, and Claude Sonnet~4.6 all reach their highest tier recall on
HIGH entities ($0.42$, $0.57$, $0.74$, and $0.77$). Thus, a single
aggregate F1 score hides the specific slice where the rule-based system
is least reliable. The Privacy Filter result should be read with one
caveat: much of its HIGH recall comes through broad ID/account labels,
while true HIGH-sensitivity categories such as religion remain difficult for
its native schema.

The disclosure-form axis shows the same split. Presidio recovers
$0.23$ of complete mentions, but only $0.02$ of partial and $0.07$ of
obfuscated mentions. GPT-4.1 and Claude Sonnet~4.6 handle partial mentions much better
($0.78$ and $0.84$ recall), though obfuscated masks remain harder.
Per-type and per-language results (Appendices~\ref{sec:pertype-app}
and~\ref{sec:perlang}) show the same ordering: LLMs lead across all
$25$ languages, Presidio is near-zero on HIGH-sensitivity and long-tail types,
and name components remain weak.

\subsection{Data Quality}
\label{sec:results-geval}\label{sec:manualval}

Two independent checks support the synthetic gold. A three-judge
G-Eval panel (GPT-5.2, Claude Sonnet~4.6,
Gemini~2.5~Pro) gives a cross-judge weighted mean of $8.04/10$
(Table~\ref{tab:geval}); we treat this as a diagnostic check rather
than ground truth. Two annotators manually validated $510$ entities
($10$ per type) for type, span, disclosed flag, disclosure form, and
sensitivity tier. Agreement with the gold is $99.1\%$ for type,
$99.1\%$ for span, $99.7\%$ for disclosed flag, and $99.8\%$ for
disclosure form; annotators also reach $96.4\%$ raw agreement with
each other. A separate $25$-record audit covering $521$ entities finds
a missed-entity rate below $3.3\%$. Both checks identify sensitivity tiering as the hardest axis: it is
the lowest-scored G-Eval dimension and the only annotation field with
substantial human disagreement ($90.5\%$ agreement, $\kappa=0.25$).
We therefore treat tier-specific findings as central but bounded by
the manual validation, explicit tiering rules, and stratified detector
results.

\begin{table}[t]
\centering
\small
\setlength{\tabcolsep}{3pt}
\begin{tabular}{l c c c}
\toprule
\textbf{Dimension} & \textbf{GPT-5.2} & \textbf{Claude} & \textbf{Gemini} \\
 & {\scriptsize $n{=}500$} & {\scriptsize $n{=}500$} & {\scriptsize $n{=}348$} \\
\midrule
Realism            & 8.74 & 8.49 & 9.28 \\
Entity validity    & 7.82 & 7.28 & 8.89 \\
Type correctness   & 7.71 & 7.60 & 8.98 \\
Disclosed flag     & 8.21 & 8.52 & 9.20 \\
\textbf{Tier correctness} & \textbf{5.77} & \textbf{6.85} & \textbf{6.81} \\
Span quality       & 8.43 & 6.94 & 9.21 \\
\bottomrule
\end{tabular}
\caption{Per-judge mean G-Eval quality scores. Tier correctness is lowest for all judges.}
\label{tab:geval}
\end{table}

\section{Discussion}
\label{sec:discussion}

The central result is that aggregate F1 is insufficient for PII
evaluation: Presidio's $0.195$ partial-overlap micro-F1 masks a
higher-risk failure, with HIGH-tier recall of only $0.07$ versus
$0.74$ for GPT-4.1 and $0.77$ for Claude Sonnet~4.6. Disclosure form
shows the same pattern: Presidio recalls only $0.02$ of partial and
$0.07$ of obfuscated mentions, while LLM detectors handle partial
mentions more reliably. These slices matter because HIGH-sensitivity
attributes and non-verbatim identifiers are precisely where aggregate
metrics can make a detector look safer than it is. Because sensitivity
tiering is also difficult ($90.5\%$ agreement, $\kappa=0.25$, and the
lowest G-Eval scores), future PII systems need both span detection \& reliable risk-aware classification.

\section{Conclusion}
\label{sec:conclusion}

We presented \textsc{REDACT}, a systematically controlled multilingual
PII benchmark with $13{,}427$ records, $324{,}078$ entity annotations,
$51$ entity types, and $25$ languages. Its nine-axis covering-array
design and entity-level metadata for disclosed status, disclosure form,
and sensitivity tier enable stratified evaluation beyond aggregate F1.
On a locked $1{,}000$-record sample, Claude Sonnet~4.6 leads
partial-overlap micro-F1 ($0.636$), followed by GPT-4.1 ($0.597$),
while the stratified results show that rule-based detection is weakest
on HIGH-sensitivity and non-verbatim PII. We release the benchmark,
schema, prompts, and harness to support more compliance-aware audits of
PII detection systems.

\section*{Limitations}
\label{sec:limitations}

We describe five limitations of the present work and the follow-on work needed to address them.

\paragraph{Synthetic--real validity is not yet quantified.}
The benchmark includes infrastructure for comparing detector rankings
between synthetic and real PII data, but this paper does not report a
validated correlation coefficient. The main obstacle is data coverage:
public real-data PII corpora are much narrower than this benchmark in
language, domain, and entity-type coverage. TAB \citep{pilan2022tab},
for example, is English, legal-domain data with four NER-style
categories. A correlation on that small overlap would test only that
slice, not the full benchmark. The i2b2/n2c2 clinical corpora
\citep{stubbs2017identifying} are a natural next comparison, but they
require a data-use agreement whose timeline did not permit inclusion.

\paragraph{Paired-sweep axis-effect analysis is deferred.}
The sampler supports a content-controlled paired-sweep mode
(Section~\ref{sec:sampler}) for isolating the effect of one axis at a
time, but this paper does not report paired-sweep deltas. That analysis
requires a separate controlled generation and detector run, so we
release the mode and leave the quantitative axis-effect results to
future work.

\paragraph{Detector experiments use sampled records.}
All five detectors are evaluated on a locked, language-stratified
sample of $1{,}000$ records from the full $13{,}427$-record corpus,
mainly because frontier-model inference is costly. The sample is sized
for aggregate and per-language comparisons, but estimates for the
rarest entity types have wider intervals. We therefore report
confidence intervals for headline comparisons.

\paragraph{HIGH-sensitivity results may depend on synthetic wording.}
The LLM detectors' strong recall on HIGH-sensitivity entities may partly
reflect how clearly those categories are expressed in generated text.
Religion, health, and orientation mentions often appear as distinctive
content words, whereas identifiers such as account numbers may be
embedded in strings. We therefore treat the HIGH-sensitivity inversion as a
finding about this benchmark and detector set, especially Presidio's
weakness on this slice, rather than as a universal claim about all
real-world HIGH-sensitivity data.

\paragraph{Triple-name decomposition remains difficult.}
The one release gate the corpus does not fully clear is B4, which
checks whether multi-component personal names are decomposed into their
constituent mentions. Compliance is high for given-name and surname
structures, but it degrades for naming systems with patronymics,
compound surnames, or non-positional ordering. Locale-aware name
decomposition is therefore a priority for the next catalog revision.

\section*{Ethical Considerations}
The benchmark contains no real personal data. Every identifier is
fabricated by the generator under an explicit fictional-content rule
(R3, Appendix~\ref{sec:genprompt}): values are structurally plausible
and satisfy real checksum and format specifications so that detectors
behave realistically, but they correspond to no real individual. The
resource is intended to evaluate and improve PII detection in support
of data-protection compliance. We document where detectors fail, for
example the rule-based weakness on HIGH-sensitivity categories, to
prevent unsafe deployment rather than to enable evasion.

The released pattern catalog includes realistic surface forms and
checksum-valid formats because such structure is necessary to test PII
detectors. To reduce misuse risk, the release will include
acceptable-use terms restricting the resource to research, evaluation,
auditing, and defensive privacy work; it will prohibit attempts to use
the catalog for impersonation, credential generation, evasion, or
re-identification. The corpus itself exposes no real PII, but
deployment of detectors trained or evaluated with it remains the
responsibility of practitioners in their own legal jurisdictions. We
release the corpus, schema, and harness for research use.



\bibliography{custom}

\appendix

\section{Pattern Catalog Construction}
\label{sec:catalog-construction}
The $4{,}127$-pattern catalog (Section~\ref{sec:catalog}) was compiled
to maximize coverage of the surface forms each entity type takes
across the $25$ languages. For every (entity type $\times$ language
$\times$ surface-form class) cell, four annotators independently
enumerated candidate patterns with the assistance of a large reasoning
model (Claude Opus~4.6), each prompted to surface rare and long-tail
realizations alongside common ones; we then retained the \emph{union}
of the four enumerations, so any pattern produced by any annotator is
kept. Independent enumeration followed by union reduces the blind
spots of any single prompt or annotator and is the basis for our
coverage claim: the catalog is \emph{broad} rather than exhaustive,
since no finite catalog can enumerate every surface form a PII entity
may take. Pattern validity is grounded in model-agnostic format
specifications rather than in the enumerating model: country-specific
identifiers satisfy real checksum schemes (Luhn, IBAN mod-97,
national-ID), dates use the appropriate calendar system, and names
follow locale-appropriate distributions. A pattern is therefore correct
independently of how it was surfaced. The enumeration prompts, the
per-annotator outputs, and the merged catalog are released with the
benchmark.

\section{Release Gates}
\label{sec:gates}
The release-gate protocol of Section~\ref{sec:qa} comprises six
critical metrics. Five target annotation integrity: B1 (offset
accuracy $=100\%$), B2 (zero-entity rate $\leq 0.5\%$), B4
(triple-name decomposition), B5 (canonical-type compliance
$=100\%$), and B6 (multi-mention completeness $\geq 95\%$). One gate
targets coverage, A2 (per-type record floor $\geq 50$).

\section{Sample Sizes and Statistical Power}
\label{sec:power}
Per-language resolution is the binding constraint on the
$1{,}000$-record detector sample, which sits well above the Cochran
floor for a $\pm5\%$ margin \citep{cochran1977sampling}.
Manual-validation entities are judged independently by two
annotators. The released evaluation harness supports confidence-interval estimation for headline comparisons; in this paper, we treat the 1{,}000-record detector run as a fixed, language-stratified evaluation sample.

\section{Generation Prompt}
\label{sec:genprompt}
The Stage-2 generator is governed by a single structured prompt: a
system message encoding the hard rules below, a user message carrying
the fully resolved nine-axis configuration, and ten few-shot anchors
covering the principal generation conditions (tight co-occurrence
clusters, hypothetical/negated mentions, partial and obfuscated
forms, OCR distortion, heavy code-switching, non-Latin scripts, and
the paired-sweep mode). The model emits the passage and the entity
list only; character offsets are computed downstream
(Section~\ref{sec:pipeline}). We summarize the rules here; the full
prompt and all anchors are released with the benchmark.

\begin{description}[leftmargin=2.4em,itemsep=1pt,topsep=2pt,font=\normalfont\ttfamily]
\item[R1.] Emit passage and entity strings only; no character
offsets (computed downstream by deterministic alignment).
\item[R2.] Every listed entity string appears \emph{verbatim} in the
passage (same casing, diacritics, script, whitespace).
\item[R3.] All PII is fictional: plausible structure, fabricated
content; no real people, numbers, or addresses.
\item[R4.] One list entry \emph{per textual occurrence}, not per
distinct entity; repeated mentions each get an entry with an
incrementing \texttt{mention\_index}.
\item[R5.] The passage satisfies all nine axis constraints at once;
on apparent conflict the \texttt{language} axis is the matrix and
\texttt{code\_switching} the embedded language.
\item[R6.] Adjacency (\texttt{none}/\texttt{loose}/\texttt{tight}) is
the strongest structural constraint; \texttt{tight} follows the named
co-occurrence (CO-\#\#\#) template.
\item[R7.] Each entity carries three semantic fields:
\texttt{disclosed} (real disclosure vs.\ hypothetical/reported),
\texttt{disclosure\_form} (\texttt{complete}/\texttt{partial}/%
\texttt{obfuscated}), and \texttt{mention\_index}.
\item[R8.] OCR distortion uses realistic substitutions
(l$\leftrightarrow$1$\leftrightarrow$I, O$\leftrightarrow$0,
rn$\leftrightarrow$m, dropped diacritics, spacing and ligature
errors); distorted strings are flagged \texttt{obfuscated}.
\item[R9.] Code-switching is grammatical mixing within one passage
(light $=$ borrowings, heavy $=$ phrase/sentence alternation, none
$=$ single language), not translation.
\item[R10.] A pre-emission self-check verifies R2--R9 and R11--R16:
verbatim presence, adjacency, occurrence counts, target-pattern
presence, density-tier match, \texttt{disclosed} flags on
hypothetical frames, triple-name decomposition, format validity,
canonical-type membership, and disclosure-form consistency.
\item[R11.] Triple-name rule: every \texttt{Full\_Name} also emits
nested \texttt{First\_Given\_Name} and \texttt{Last\_Family\_Name}
spans, with locale-aware decomposition (Western, CJK, Hungarian,
Spanish double-surname, Russian patronymic; particles kept with the
surname, honorifics excluded).
\item[R12.] Format-validity contract: the passage is machine-parseable
in the requested format and generated identifiers satisfy their
real-world checksum/format specs (Luhn, IBAN mod-97, country
national-ID schemes).
\item[R13.] All entity types are drawn from the $51$ canonical types
only.
\item[R14.] Sensitivity-tier and HIGH-sensitivity awareness: each entity
is tiered \texttt{HIGH}/\texttt{MEDIUM}/\texttt{LOW} by regulatory
sensitivity.
\item[R15.] Script-aware definition of inter-entity spacing for
non-whitespace-delimited scripts (CJK, Thai, Arabic), used by the
adjacency check.
\item[R16.] Frozen-entity pinning (paired-sweep mode only): supplied
frozen entity values appear verbatim and are present in the output
entity list.
\end{description}

\section{Sensitivity-Tier Decision Rules}
\label{sec:tier-rules}
Sensitivity tiers are assigned from the entity type and the disclosure
context, not from detector confidence. Table~\ref{tab:tier-rules}
summarizes the operational rules used by the generator, judge prompt,
and manual validation.

\begin{table*}[t]
\centering
\scriptsize
\setlength{\tabcolsep}{4pt}
\renewcommand{\arraystretch}{1.14}
\begin{tabular}{p{0.16\textwidth} p{0.28\textwidth} p{0.25\textwidth} p{0.22\textwidth}}
\toprule
\textbf{Tier} & \textbf{Included cases} & \textbf{Boundary rule} & \textbf{Examples} \\
\midrule
HIGH & Article~9-style special categories in the taxonomy: Religion, Sex\_Orientation, Political\_Party, Trade\_Union\_Membership, Crime, Medical\_Information, Allergy\_Information, and Ethnicity. Date\_of\_Birth is also marked HIGH in this benchmark because of identity-linkage risk. & Use HIGH only when the entity type itself belongs to this set. If the text merely names a category without disclosing a person's attribute, keep the type/tier but set \texttt{disclosed=false}. & ``Religion: Catholic'', ``allergy to penicillin'', ``DOB 12.05.1998''. \\
MEDIUM & Direct identifiers that can identify or contact a person but are not HIGH-sensitivity-style categories: names, emails, phone numbers, personal and work addresses, government IDs, account numbers, passport numbers, driving-license numbers, tax references, credit-card numbers, IP addresses, passwords, and social-media identifiers. & Prefer MEDIUM for values that directly identify an individual, account, or credential. Contact information and IDs are not promoted to HIGH unless the entity type itself is a HIGH category. & ``J. Smith'', ``arya@example.com'', ``passport K1234567'', ``ending 8473''. \\
LOW & Quasi-identifiers and contextual attributes: city, state, country of residence, place of birth, nationality, business title, gender, marital status, citizenship status, organization name, professional background, and similar contextual fields. & Keep LOW when the value is identifying only in combination with other fields. Nationality and country of residence are LOW unless the annotation is explicitly Ethnicity or another HIGH category. & ``Hyderabad'', ``Product Manager'', ``country of residence: Japan''. \\
Disclosure boundary & Applies across all tiers. & Hypotheticals, examples, templates, empty form labels, and category mentions are annotated with \texttt{disclosed=false}; actual attributes of an identifiable person are \texttt{disclosed=true}. & ``Religion'' in a hate-speech policy note is not a personal disclosure; ``Religion: Hindu'' in an employee record is. \\
Medical boundary & Usually HIGH when the text states a person's condition, treatment, diagnosis, allergy, or clinical status. & General health-related words are not HIGH disclosures unless they describe a specific person's medical attribute. & ``diagnosed with asthma'' is HIGH and disclosed; ``medical information required'' is not a disclosed medical attribute. \\
\bottomrule
\end{tabular}
\caption{Operational sensitivity-tier rules used for annotation, judging, and manual validation. The tiers are benchmark labels for evaluation slices; they should not be read as legal advice.}
\label{tab:tier-rules}
\end{table*}

\section{Quality-Assessment Rubric}
\label{sec:rubric}
The reference-free LLM-as-judge protocol (Section~\ref{sec:geval})
scores each record on a $1$--$10$ scale along six dimensions, each
with four ordinal anchor levels. The six dimensions and weights are:
realism ($2.0$), entity validity ($2.0$), type correctness ($1.5$),
disclosed-flag correctness ($1.5$), tier correctness ($1.0$), and
span quality ($1.0$). Table~\ref{tab:rubric-anchors} gives the anchor
ranges used in the judge prompt. The prompt also instructs judges to
score each dimension independently, avoid penalizing synthetic text
only because it is synthetic, avoid rewarding annotation volume, and
apply the same HIGH-sensitivity / direct identifier / quasi-identifier tier
mapping for all records.

\begin{table*}[t]
\centering
\scriptsize
\setlength{\tabcolsep}{3pt}
\renewcommand{\arraystretch}{1.12}
\begin{tabular}{p{0.13\textwidth} p{0.20\textwidth} p{0.20\textwidth} p{0.20\textwidth} p{0.20\textwidth}}
\toprule
\textbf{Dimension} & \textbf{$1$--$2$} & \textbf{$3$--$4$} & \textbf{$5$--$7$} & \textbf{$8$--$10$} \\
\midrule
Realism & Nonsensical, contradictory, or obviously fake. & Clearly AI-generated, template-like, repetitive, or anachronistic. & Mostly plausible, with minor synthetic phrasing or mild inconsistencies. & Natural, domain-appropriate, internally consistent, and close to authentic real-world text. \\
Entity validity & Most annotations are not actual PII about a specific person. & Many annotations are abstract categories rather than personal disclosures. & Most annotations are genuine, but a few are category mentions, templates, or examples. & Annotated spans are genuine personal disclosures about identifiable individuals. \\
Type correctness & Most entity-type labels are wrong. & Multiple labels are wrong, including obvious type errors. & Most labels are correct, with a few boundary errors between similar types. & Labels match the $51$-type taxonomy consistently. \\
Disclosed flag & The field appears arbitrary and does not follow the text. & Multiple category mentions are marked disclosed, or genuine disclosures are missed. & Mostly correct, with a small number of clear mistakes. & Correctly separates actual disclosures from examples, templates, category mentions, and hypotheticals. \\
Tier correctness & Tier assignments appear arbitrary. & Systematic errors, such as HIGH-sensitivity categories at MEDIUM or direct identifiers at HIGH. & Mostly correct, with some over-tiering or under-tiering at boundaries. & Tiers follow the fixed HIGH-sensitivity / direct identifier / quasi-identifier mapping. \\
Span quality & Spans are systematically malformed. & Multiple spans contain substantial extra context or merge multiple entities. & Some spans include extra context but still identify the entity. & Spans are cleanly bounded and contain exactly the identifier. \\
\bottomrule
\end{tabular}
\caption{Four-level rubric anchors used by the LLM-as-judge quality panel. Judges assign scores from $1$ to $10$ within each dimension, using the anchor ranges as calibration points.}
\label{tab:rubric-anchors}
\end{table*}

\section{Sampler Axis Values}
\label{sec:axes}
The covering array of Section~\ref{sec:sampler} is constructed over
the full value set of the nine axes, enumerated below so the
sampling process is reproducible.

\begin{description}[leftmargin=1.2em,itemsep=1pt,topsep=2pt]
\item[\texttt{domain} (12):] healthcare, legal, finance, education,
government, technology, HR, insurance, customer\_service,
law\_enforcement, social\_media, e\_commerce.
\item[\texttt{format} (7):] plain\_text, email, chat\_transcript,
ticket\_worknotes, json\_record, key\_value\_pairs, log\_entry.
\item[\texttt{difficulty} (3):] easy, medium, hard (extraction hardness).
\item[\texttt{length} (4):] small ($50$--$200$ tokens), medium
($200$--$600$), large ($600$--$2{,}000$), very\_large
($2{,}000$--$4{,}000$).
\item[\texttt{density} (3):] low ($1$--$4$ entities / $1$k tokens),
medium ($5$--$15$), high ($\geq 16$).
\item[\texttt{code\_switching} (3):] none, light, heavy (grammatical
language mixing within a record, not translation).
\item[\texttt{language} (25):] AR (Arabic), CS (Czech), DA (Danish),
DE (German), EN (English), ES (Spanish), FI (Finnish), FR (French),
FR\_CA (Canadian French), HE (Hebrew), HI (Hindi), HU (Hungarian),
IT (Italian), JA (Japanese), KO (Korean), NL (Dutch), NO (Norwegian),
PT\_BR (Brazilian Portuguese), PT\_EU (European Portuguese),
RU (Russian), SV (Swedish), TH (Thai), TR (Turkish),
ZH\_CN (Simplified Chinese), ZH\_TW (Traditional Chinese).
\item[\texttt{adjacency} (3):] none (entities $\geq 10$ words apart),
loose ($3$--$10$ words), tight (a co-occurrence cluster).
\item[\texttt{co\_occurrence\_pattern} (15 + sentinel):] CO-001
through CO-015, each a named dense-cluster template, e.g.\ CO-008
angle-bracket email metadata, CO-009 patient header, CO-010
authentication cluster, CO-011 compliance cluster, CO-012 HR record,
CO-013 contact card, CO-014 legal citation, CO-015 form key-value
adjacency, plus a \texttt{none} sentinel when \texttt{adjacency}
$\neq$ tight.
\end{description}

\section{Benchmark Comparison}
\label{sec:relatedtable}
Table~\ref{tab:related} positions the present work against prior PII
benchmarks along seven dimensions.

\begin{table*}[t]
\centering
\small
\setlength{\tabcolsep}{2pt}
\begin{tabular}{l c c c c c c c}
\toprule
\textbf{Benchmark} & \textbf{\#Types} & \textbf{\#Lang.} & \textbf{Public} & \textbf{Synthetic} & \textbf{HIGH slice} & \textbf{Stratified eval.} & \textbf{Behavioral frames} \\
\midrule
i2b2/n2c2 \citep{stubbs2017identifying}      & 25 & 1  & DUA       & no        & no   & no            & no \\
TAB \citep{pilan2022tab}                      & 10 & 1  & yes       & no        & no   & no            & no \\
SPY \citep{savkin2025spy}                     & 7  & 1  & yes       & yes       & no   & no            & no \\
AI4Privacy 1m \citep{ai4privacy2025openpii}   & 19 & 23 & yes       & templated & no   & no            & no \\
PIIBench \citep{jha2026piibench}              & 48 & 1  & yes       & consol.   & no   & per-type only & no \\
MultiCoNER~II \citep{fetahu2023multiconer2}   & 33 & 12 & yes       & partial   & n/a  & no            & no \\
\midrule
\textbf{This work}                            & \textbf{51} & \textbf{25} & \textbf{yes} & \textbf{yes} & \textbf{yes} & \textbf{F5, F6} & \textbf{7 frames} \\
\bottomrule
\end{tabular}
\caption{Comparison with prior PII benchmarks. The present work is
the first to combine $25$-language coverage, explicit GDPR
HIGH-sensitivity alignment, and a metadata-driven stratified-evaluation
framework. ``Stratified eval.'' indicates evaluation slices beyond
per-entity-type F1. ``DUA'' = data-use agreement required;
``consol.'' = consolidated from existing sources.}
\label{tab:related}
\end{table*}

\section{Per-Language Results}
\label{sec:perlang}
\begin{figure*}[t]
\centering
\includegraphics[width=0.98\textwidth]{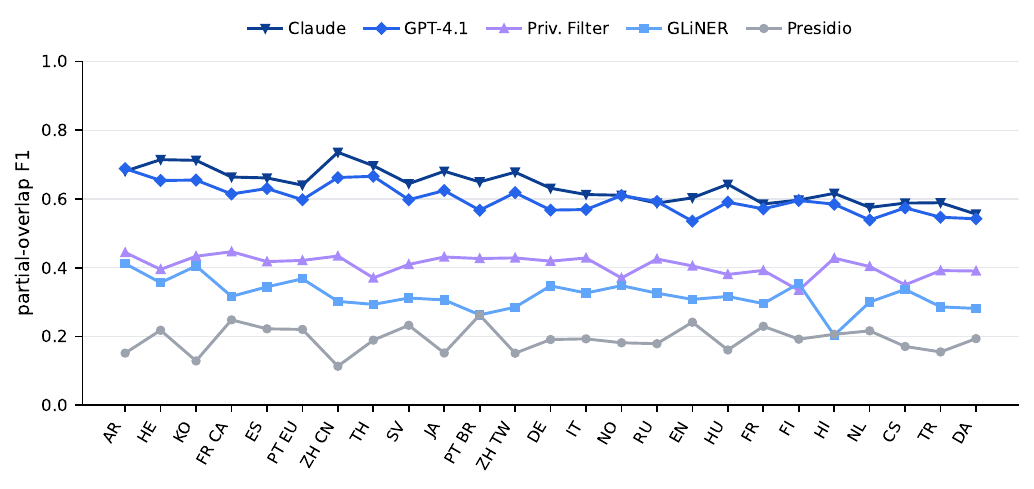}
\caption{Per-language partial-overlap F1 for each detector across all
$25$ languages (sorted by mean F1). The two LLM detectors lead and
Presidio trails on every language; the middle tier (GLiNER, Priv.\
Filter) runs close and swaps on a few languages. The stable top and
bottom confirm that the detector-family ranking holds across the
multilingual corpus.}
\label{fig:perlang}
\end{figure*}

Figure~\ref{fig:perlang} reports partial-overlap F1 for each detector
across all $25$ languages, sorted by mean F1.

\section{Anatomy of a Generated Record}
\label{sec:anatomy}
\begin{figure*}[t]
\centering
\includegraphics[width=0.92\textwidth]{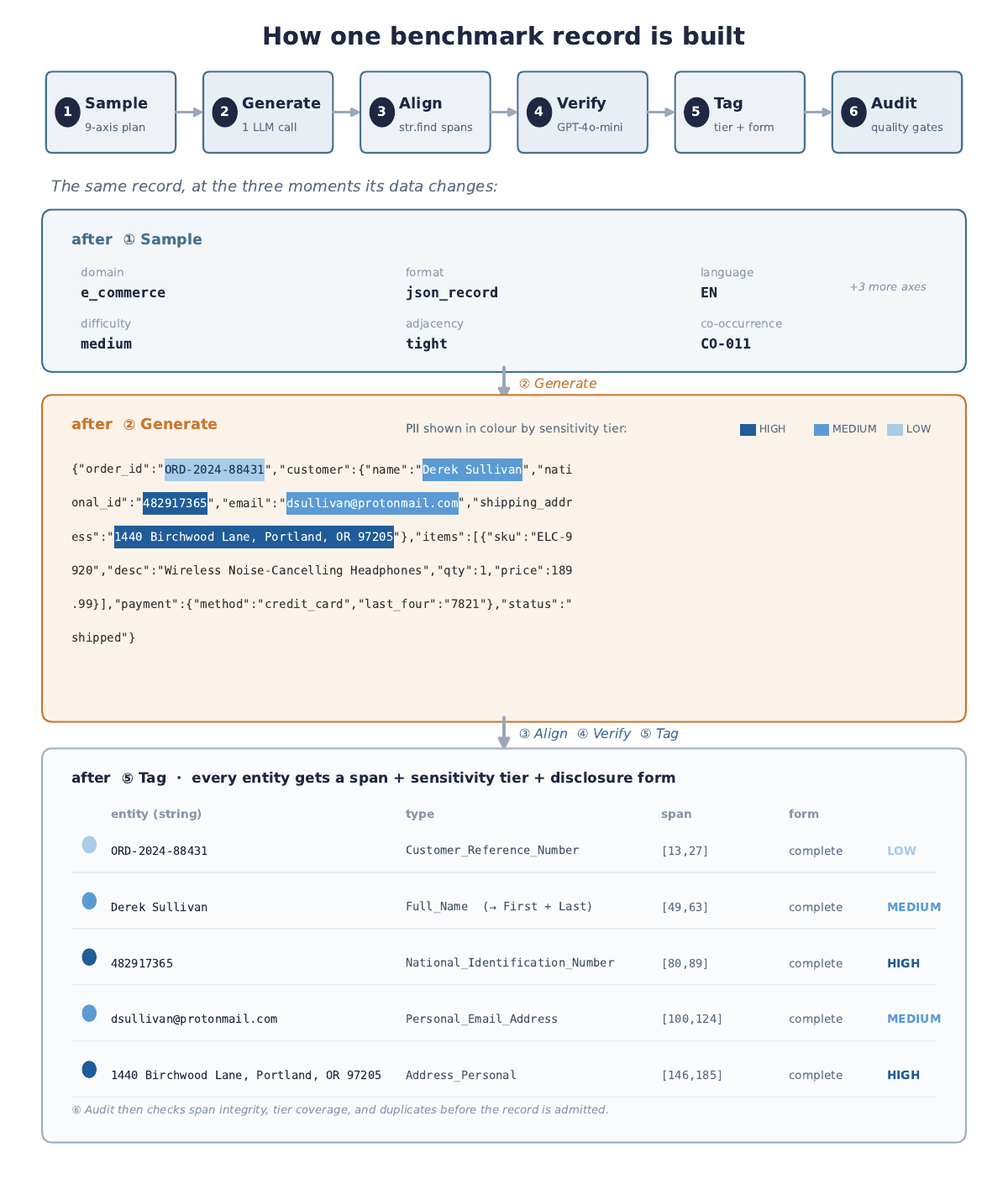}
\caption{How one benchmark record is built. The figure traces a
single generated record from its nine-axis sampler plan to the
generated text and final metadata-tagged annotations. The example
illustrates how planned generation conditions are converted into a
record whose entity mentions receive type, span, disclosure-form, and
sensitivity-tier annotations before release.}
\label{fig:record-example}
\label{fig:anatomy}
\end{figure*}

Figure~\ref{fig:anatomy} traces a single record from its sampler
plan row through generation to the aligned, metadata-tagged
annotations, on a real example drawn from the corpus.

\section{Per-Type Results}
\label{sec:pertype-app}
\begin{figure}[t]
\centering
\includegraphics[width=\columnwidth]{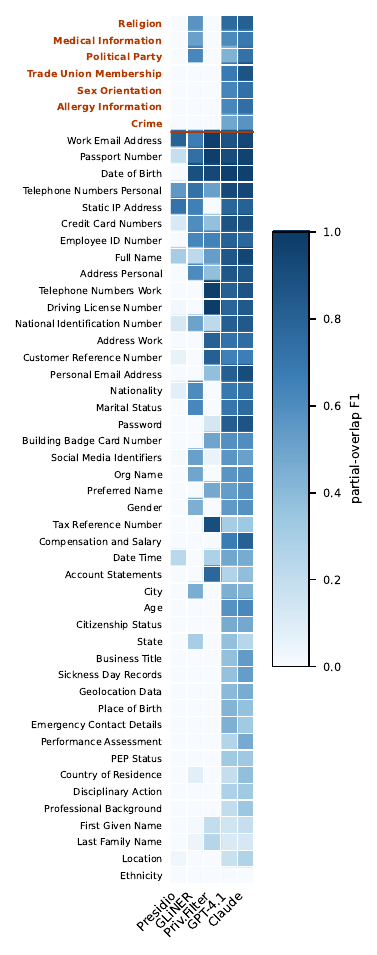}
\caption{Per-type partial-overlap F1 for each detector, with the
GDPR Article~9 types banded at top. Presidio (leftmost column) is
near-zero on Article~9 and on most types; the LLM detectors cover
the long tail. Name-component types are weak across all detectors.}
\label{fig:pertype}
\end{figure}

Figure~\ref{fig:pertype} reports partial-overlap F1 for each detector
on each of the $51$ canonical types, with the GDPR HIGH-sensitivity types
banded at top.

\end{document}